\documentclass[letterpaper, 10 pt, conference]{ieeeconf}  

\IEEEoverridecommandlockouts                              

\title{\LARGE \bf
CSC-MPPI: A Novel Constrained MPPI Framework \\ with DBSCAN for Reliable Obstacle Avoidance*
}

\author{Leesai Park$^{1}$, Keunwoo Jang$^{2, \dagger}$, and Sanghyun Kim$^{1, 3, \dagger}$
\thanks{*This research was supported by the National Research Foundation of Korea (NRF) grant funded by the Korea government (MSIT) (No. RS-2024-00461583 and No. RS-2024-00411007). Also, this work was supported by KIST Institutional Program (No. 2E33591, 2I25530) and Hyundai Motor Company and Kia.}
\thanks{$^{1}$L. Park and S. Kim are with the Department of Mechanical Engineering, Kyung Hee University, Yongin, Republic of Korea {\tt\small \{leesai2000, kim87\}@khu.ac.kr}}%
\thanks{$^{2}$K. Jang is with the Center for Humanoid Research, Korea Institute of Science and Technology, Seoul, Republic of Korea {\tt\small jang90@kist.re.kr}}%
\thanks{$^{3}$S. Kim is also with Advanced Institutes of Convergence Technology
 (AICT), Suwon, Republic of Korea }%
\thanks{$^{\dagger}$Corresponding Authors: Keunwoo Jang and Sanghyun Kim.}
}
\usepackage{cite}
\usepackage{bm}
\usepackage{amsmath}
\usepackage{amsfonts}

\usepackage{graphicx}
\usepackage{subfig}
\usepackage{hyperref}
\usepackage{algorithm}
\usepackage{algpseudocode}
\usepackage{mathtools}
\usepackage{verbatim}
\usepackage{booktabs}
\usepackage{makecell}
\usepackage{xcolor}
\begin{document}

\maketitle
\thispagestyle{empty}
\pagestyle{empty}

\begin{abstract}

This paper proposes Constrained Sampling Cluster Model Predictive Path Integral (CSC-MPPI), a novel constrained formulation of MPPI designed to enhance trajectory optimization while enforcing strict constraints on system states and control inputs. Traditional MPPI, which relies on a probabilistic sampling process, often struggles with constraint satisfaction and generates suboptimal trajectories due to the weighted averaging of sampled trajectories. To address these limitations, the proposed framework integrates a primal-dual gradient-based approach and Density-Based Spatial Clustering of Applications with Noise (DBSCAN) to steer sampled input trajectories into feasible regions while mitigating risks associated with weighted averaging. First, to ensure that sampled trajectories remain within the feasible region, the primal-dual gradient method is applied to iteratively shift sampled inputs while enforcing state and control constraints. Then, DBSCAN groups the sampled trajectories, enabling the selection of representative control inputs within each cluster. Finally, among the representative control inputs, the one with the lowest cost is chosen as the optimal action. As a result, CSC-MPPI guarantees constraint satisfaction, improves trajectory selection, and enhances robustness in complex environments. Simulation and real-world experiments demonstrate that CSC-MPPI outperforms traditional MPPI in obstacle avoidance, achieving improved reliability and efficiency. The experimental videos are available at \hyperlink{URL}{https://cscmppi.github.io}
\end{abstract}

\begin{figure}[t]
    \centering
    \subfloat[]{
        \includegraphics[width=0.40\linewidth]{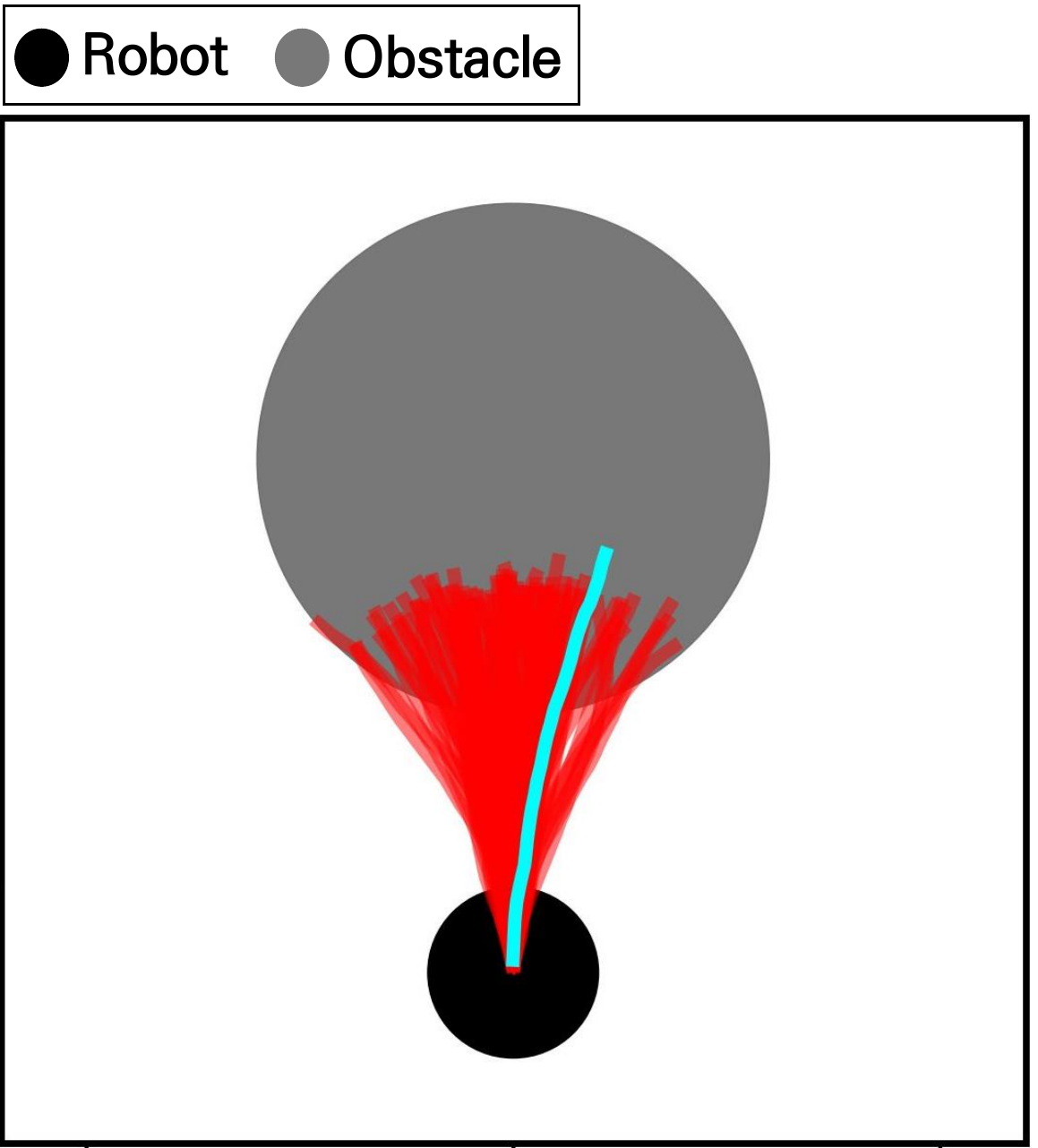} 
        \label{fig1a}
    }
    \subfloat[]{
        \includegraphics[width=0.40\linewidth]{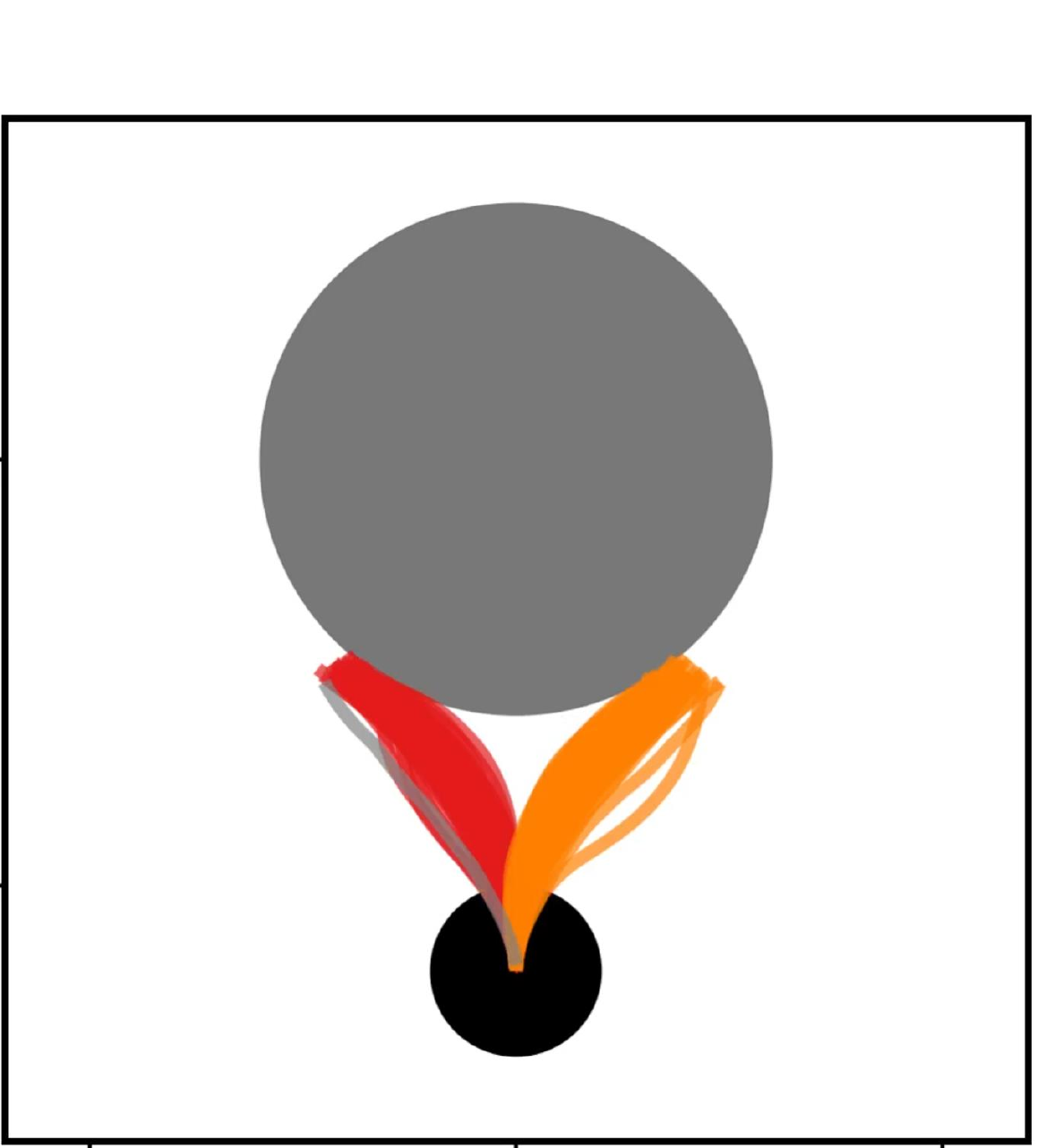} 
        \label{fig1b}
    }
    \vspace{-3mm}
    \subfloat[]{
            \includegraphics[width=0.40\linewidth]{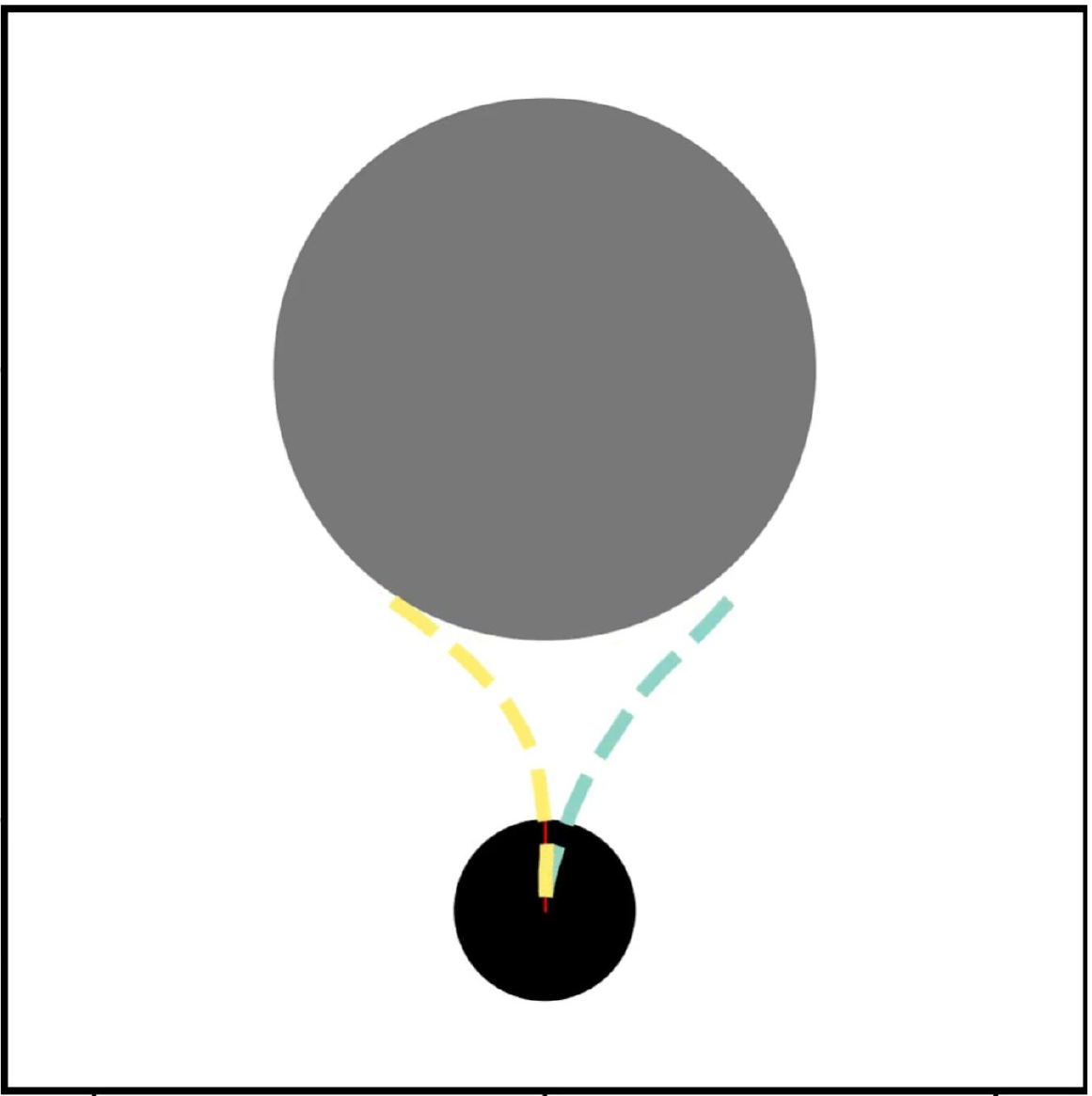} 
            \label{fig1c}
        }
        \subfloat[]{
            \includegraphics[width=0.40\linewidth]{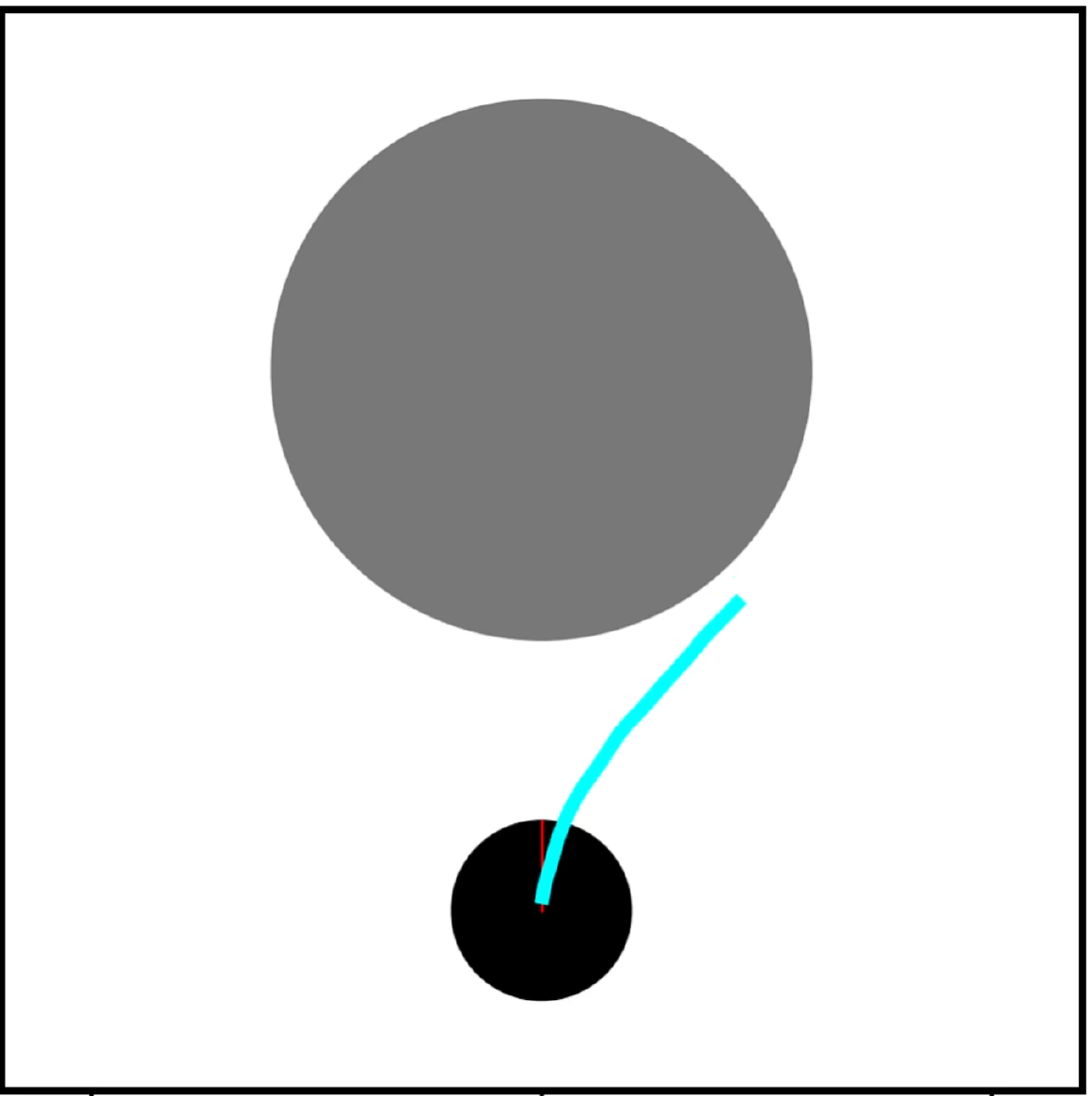} 
            \label{fig1d}
        }

    \caption{Comparison of standard MPPI and CSC-MPPI. 
    (a) Sampled trajectories from standard MPPI (red solid lines) are concentrated in high-cost regions, producing a suboptimal path (cyan solid line) that results in collision.  
    (b) CSC-MPPI adjusts samples and clusters feasible trajectories (red/orange solid lines), while noise samples are shown in gray. 
    (c) A representative trajectory (light blue/yellow dashed lines) is selected from each cluster.
    (d) The final optimal trajectory (cyan solid line) is chosen as the lowest-cost candidate.}
    \label{fig1}
    \vspace{-6mm}
\end{figure}


\section{INTRODUCTION} \label{sec1}

Model Predictive Path Integral (MPPI) \cite{williams2016aggressive} is a sampling-based Model Predictive Control (MPC) method that optimizes control inputs by evaluating a large number of trajectory samples drawn from a stochastic distribution. Unlike traditional MPC frameworks \cite{wang2024online,incremona2017mpc,faulwasser2016implementation, todorov2005generalized}, which typically require iterative optimization, MPPI directly approximates the optimal control distribution through sampling, making it highly effective for real-time trajectory planning in dynamic and uncertain environments. By incorporating stochasticity into the control process, MPPI enhances robustness and flexibility, allowing it to adapt to rapidly changing conditions.

Despite its advantages, MPPI still faces several limitations, particularly in constraint satisfaction. Because MPPI relies on a weighted averaging process to compute the final control input, it often struggles to enforce hard constraints on state and control variables. Thus, as shown in Fig. \ref{fig1a}, when sampled trajectories are overly concentrated in high-cost regions or lack sufficient diversity, MPPI becomes susceptible to local minima and potential collisions with obstacles.

Hence, in this paper, we propose a novel MPPI framework—Constrained Sampling Cluster MPPI (CSC-MPPI)—which integrates a primal-dual gradient-based adjustment with Density-Based Spatial Clustering
of Applications with Noise (DBSCAN) clustering to efficiently enforce control constraints and ensure trajectory feasibility.

\subsection{Related Works} \label{sec11}

In traditional MPC frameworks, extensive research has been conducted to incorporate constraints into optimization methods such as Sequential Quadratic Programming (SQP) and Differential Dynamic Programming (DDP), leading to improved control performance in complex environments \cite{marti2020squash, tassa2014control, xie2017differential, ghaemi2009integrated}. However, few studies have focused on constraint enforcement within MPPI as its inherent reliance on random sampling makes it challenging to impose strict constraints. This limitation restricts applicability in safety-critical or constrained environments.

To address this problem, Yan and Devasia \cite{yan2024output} introduced Output-Sampled MPPI (o-MPPI), which improves constraint satisfaction by selectively sampling within predefined acceptable regions in the output space. However, o-MPPI does not provide a clear strategy for handling samples that fall outside these acceptable regions. Balci \textit{et al.} \cite{balci2022constrained} proposed Constrained Covariance Steering-Based MPPI (CCS-MPPI), which enforces state and input constraints through hyperplane-based method. Nonetheless, its reliance on probabilistic constraints can lead to occasional violations. To strengthen constraint handling in MPPI, Yin \textit{et al.} \cite{yin2023shield} developed Shield-MPPI, which combines a dual-layer safety mechanism with Control Barrier Functions (CBFs), a cost function that penalizes unsafe trajectories, and a reactive safety shield that adjusts controls in real time. Although this structured approach enhances safety, it still depends on soft constraints and cannot fully guarantee constraint satisfaction. Finally, Tao \textit{et al.} \cite{tao2022path} introduced Stochastic Control Barrier Function MPPI (SCBF-MPPI), which embeds Stochastic Control Barrier Functions (SCBFs) directly into the sampling process, dynamically shifting the mean and covariance of the sampling distribution to ensure trajectories remain within safe regions. However, SCBF-MPPI also operates probabilistically, thus absolute constraint enforcement remains unattainable.

In summary, while these approaches each enhance MPPI’s constraint-handling capabilities, they all share a common reliance on soft constraints. Consequently, they cannot strictly guarantee state and control constraint satisfaction.

\subsection{Overview of Our Approach} \label{sec12}
To overcome the limitations of conventional MPPI in constraint enforcement, we propose CSC-MPPI, which integrates explicit hard constraint satisfaction into the MPPI framework while preserving its adaptability. The proposed approach refines the standard MPPI sampling strategy through a sequential process to ensure feasibility under strict state and input constraints.

 Initially, trajectories are sampled from a Gaussian proposal distribution centered around the initial estimate control input, following the standard MPPI formulation. To enforce constraints, each sampled trajectory undergoes an iterative adjustment process using a primal-dual gradient-based method \cite{du2019linear}, which shifts infeasible trajectories into the feasible region while maintaining their structural integrity. This step ensures that the sampled trajectories comply with both state and control constraints. Following constraint enforcement, the adjusted trajectories are grouped based on spatial proximity and cost similarity using DBSCAN \cite{deng2020dbscan} (see Fig. \ref{fig1b}). The clustering process mitigates the effects of weighted averaging by filtering out outliers and preventing high-cost trajectories from influencing control selection.  Finally, from each identified cluster, a representative control input is selected (see Fig. \ref{fig1c}), and among these candidates, the trajectory with the lowest cost is chosen as the final control action (see Fig. \ref{fig1d}). Consequently, the proposed MPPI framework can generate optimal trajectories while strictly satisfying hard constraints.
 

The main contributions of this paper are as follows:
\begin{itemize}
    \item \textbf{Hard constraint enforcement:} Unlike conventional MPPI methods that rely on soft penalties, CSC-MPPI explicitly enforces hard constraints on both state and control inputs. To the best of the authors' knowledge, this is the first MPPI-based approach to guarantee hard constraint satisfaction in both domains.
    
    \item \textbf{Robust control input selection via clustering:} By employing DBSCAN, CSC-MPPI prevents high-cost or noisy trajectories from affecting control selection. This addresses the critical limitation of weighted averaging in standard MPPI, which often leads to constraint violations.

    \item \textbf{Improved safety and efficiency:} The combination of constraint-aware sampling and clustering leads to trajectories that are not only feasible but also more efficient, reducing the risk of local minima. This has been consistently validated in both simulation and real-world navigation experiments.
\end{itemize}

The remainder of this paper is organized as follows. Section~\ref{sec2} reviews the MPPI algorithm and formulates the control problem. Section~\ref{sec3} presents the proposed CSC-MPPI method in detail. Section~\ref{sec4} describes experimental validations through simulations and hardware tests. Finally, conclusions are drawn in Section~\ref{sec5}.

\begin{table}[t] 
\caption{Symbol and corresponding meaning} \label{table1}
\begin{center}
\vspace{-2mm}
\begin{tabular}{cc} 
  \hline
   \multicolumn{1}{c}{Symbol}& \multicolumn{1}{c}{Description} \\
  \hline
 \hline
   $ a $ & scalar  \\
   $\boldsymbol{a} $ & vector \\
   $ \boldsymbol{A} $ & matrix  \\
   $ \underline{\boldsymbol{a}}$ and $\overline{\boldsymbol{a}}$ & lower bound and upper bound of $\boldsymbol{a}$  \\
   $ {}^k\boldsymbol{a}_{i} $ & $i$-th vector of the $k$-th sampled sequence \\
   $ {}^k\boldsymbol{A} $ & $k$-th sampled sequence \\
   \hline
\end{tabular}
\end{center}
\vspace{-6mm}
\end{table}

\section{PRELIMINARIES} \label{sec2}
In this section, we provide the necessary background including problem formulation and the MPPI framework to facilitate a clear understanding of the proposed framework. To enhance readability, Table \ref{table1} shows mathematical notation and its corresponding meaning in this paper.

\subsection{Problem Formulation} \label{sec21}
Let's consider a discrete-time stochastic dynamical system,
\begin{equation} \label{eq_1}
    \boldsymbol{x}_{t+1} = f(\boldsymbol{x}_t, \boldsymbol{u}_t + \delta \boldsymbol{u}_t), 
\end{equation}
where $\boldsymbol{x}_t \in \mathbb{R}^n$, $\boldsymbol{u}_t \in \mathbb{R}^m$ are the state and the control input of the system, respectively, and $\delta \boldsymbol{u}_t \sim \mathcal{N}(0,\boldsymbol \Sigma_{\boldsymbol u})$ denotes Gaussian noise with zero mean and covariance  $\boldsymbol \Sigma_{\boldsymbol{u}}$, all at time step $t$. Given a time horizon $N$, the control sequence is defined as $\boldsymbol{U} = [\boldsymbol{u}_0, \boldsymbol {u}_1, ..., \boldsymbol u_{N-1}]^T \in \mathbb R^{N \times m}$, representing the set of control inputs over the horizon. Similarly, the state trajectory is denoted by $\boldsymbol {X} = [\boldsymbol x_0, \boldsymbol x_1, ..., \boldsymbol x_N]^T \in \mathbb R^{(N+1)\times n} $, capturing the system states across the time steps. Let $\boldsymbol x_s$ and $\boldsymbol x_f$ denote the initial state and desired state of the robot, respectively. The goal is to determine an optimal control sequence $\boldsymbol{U}^*$ that minimizes a given cost function while satisfying system constraints, guiding the robot from the initial state $\boldsymbol x_s$ to the desired state $\boldsymbol x_f$. Thus, the optimization problem can be formulated as follows:
\begin{equation} \label{eq_2}
    \begin{aligned}
        \min_{\boldsymbol{U}} \quad & J = \mathbb{E} \left[ \phi (\boldsymbol{x}_N) + \sum_{t=0}^{N-1} \left( l (\boldsymbol{x}_t) + \frac{1}{2} \boldsymbol{u}_t^T \boldsymbol R \boldsymbol{u}_t \right) \right], \\
        \text{s.t.} \quad 
        & \boldsymbol{x}_{t+1} = f \left( \boldsymbol{x}_t, \boldsymbol{u}_t + \delta \boldsymbol{u}_t \right), \quad \delta \boldsymbol{u}_t \sim \mathcal{N}(0, \boldsymbol \Sigma_u), \\
        & \mathcal{X}_{\text{rob}} (\boldsymbol{x}_t) \cap \mathcal{X}_{\text{obs}} = \emptyset, \quad \boldsymbol{g}(\boldsymbol{x}_t, \boldsymbol{u}_t) \leq 0, \\
        & \boldsymbol{x}_0 = \boldsymbol{x}_s, \quad \boldsymbol{u}_t \in \mathcal{U}, \quad \boldsymbol{x}_t \in \mathcal{X},
    \end{aligned}
\end{equation}
where $\mathcal{X}_{rob} \subset \mathcal{X}^d$ and $\mathcal{X}_{obs} \subset \mathcal{X}^d$ represent the regions occupied by the robot and the obstacles in a $d$-dimensional space, $\boldsymbol g(\boldsymbol{x}_t, \boldsymbol{{u}_t})$ denotes the inequality constraints, $\mathcal{U}$ denotes the feasible control input space, and $\mathcal{X}$ denotes the feasible state space. The cost function $J$ consists of the expectation of a running cost $l(\boldsymbol{x}_t)$, a terminal cost $\phi(\boldsymbol{x}_N)$, and the control input penalty term $\frac{1}{2} \boldsymbol{u}_t^T \boldsymbol R \boldsymbol{u}_t$, where $\boldsymbol R \in \mathbb R^{m \times m}$ is a positive-definite matrix.

\subsection{Review of the MPPI}
MPPI is a control framework that solves optimal control problems by leveraging stochastic sampling techniques. Instead of relying on traditional gradient-based methods, MPPI explores the control space by drawing samples from a predefined distribution and evaluating their performance in a cost-minimization framework. To describe the probabilistic formulation of MPPI, the probability function of the distribution $\mathbb{Q}$ can be expressed as follows:
\begin{equation} \label{eq_3}
q(\boldsymbol V) = Z^{-1} \prod_{t=0}^{N-1} \exp \left( -\frac{1}{2} (\boldsymbol{v}_t - \boldsymbol{u}_t)^T\boldsymbol  \Sigma_{\boldsymbol u}^{-1} (\boldsymbol{v}_t - \boldsymbol{u}_t) \right),
\end{equation}
where  $Z = \sqrt{(2\pi)^m |\boldsymbol \Sigma_{\boldsymbol u}|}$ is the normalization term, and  $\boldsymbol v_t = \boldsymbol u_t + \delta \boldsymbol u_t$ is the actual control input, with  $\boldsymbol v_t \sim \mathcal{N}(\boldsymbol u_t, \boldsymbol \Sigma_{\boldsymbol u})$. Although MPPI aims to determine the optimal control input sequence $\boldsymbol {U}^*$ by solving (\ref{eq_2}), finding a direct solution is challenging. Therefore, MPPI instead minimizes the Kullback-Leibler (KL) divergence between the optimal distribution $\mathbb{Q}^*$ and the proposal distribution $\mathbb{Q}$, which can be formulated as follows:
\begin{equation} \label{eq_4}
     \boldsymbol U^* \simeq \underset{\boldsymbol U}{\arg\min}\;\mathbb{D}_{\text{KL}} ( \mathbb{Q}^* \parallel \mathbb{Q}).
\end{equation}
By applying the $Free$-$energy$ and Jensen's inequality to minimize the KL divergence in (\ref{eq_4}), the optimal control formulation can be derived as follows:
\begin{equation} \label{eq_5}
    \begin{aligned}
         \boldsymbol U^* &= \underset{\boldsymbol U}{\arg\min}\; \mathbb{E}_{\mathbb{Q}^*} 
        \left[ 
        \frac{1}{2} \sum_{t=0}^{N-1} (\boldsymbol{v}_{t} - \boldsymbol{u}_{t})^T 
        \boldsymbol \Sigma_{\boldsymbol u}^{-1} (\boldsymbol{v}_{t} - \boldsymbol{u}_{t})
        \right] \\
        &= \mathbb{E}_{\mathbb{Q}^*} \left[\prescript{k}{}{\boldsymbol V}\right],
    \end{aligned}
\end{equation}
where $\prescript{k}{}{\boldsymbol V} = [\prescript{k}{}{\boldsymbol v}_0, \prescript{k}{}{\boldsymbol v}_1,...,\prescript{k}{} {\boldsymbol v}_{N-1}]^T$ represents the $k$-th control sequence sampled from the optimal distribution $\mathbb Q^*$. Since directly sampling from the optimal distribution $\mathbb Q^*$ is challenging, MPPI employs the importance sampling technique to efficiently obtain the optimal control sequence $\boldsymbol{U}^*$. The expectation with respect to the optimal distribution $\mathbb Q^{*}$ can be rewritten using importance sampling as follows:

\begin{equation} \label{eq_6}
        \boldsymbol {U}^* = 
        \int \frac{q^*(\boldsymbol V)}{q(\boldsymbol V)} 
        q(\boldsymbol V) {\boldsymbol V} \, d\boldsymbol V 
        \simeq \sum_{k=0}^{K-1} w(\prescript{k}{} {\boldsymbol V})\prescript{k}{} {\boldsymbol V},
\end{equation}

\begin{equation} \label{eq_7}
    w(\prescript{k}{}{\boldsymbol V})=\frac{1}{{\eta}} 
    \exp \left( 
    - 
    \frac{1}{\lambda}S(\prescript{k}{}{\boldsymbol V}) - \sum_{t=0}^{N-1} (\hat{\boldsymbol{u}}_t - \tilde{\boldsymbol{u}}_t)^{T} 
    \boldsymbol \Sigma_{\boldsymbol u}^{-1}\prescript{k}{}{\boldsymbol{v}}_t
    \right),
\end{equation}

\begin{equation} \label{eq_8}
    {\eta} = \sum_{k=0}^{K-1}
    \exp \left( 
    - 
    \frac{1}{\lambda}S(\prescript{k}{} {\boldsymbol V}) - \sum_{t=0}^{N-1} (\hat{\boldsymbol{u}}_t - \tilde{\boldsymbol{u}}_t)^T
    \boldsymbol \Sigma_{\boldsymbol u}^{-1} \prescript{k}{}{\boldsymbol{v}}_t 
    \right),
\end{equation}
where $S(\prescript{k}{}{\boldsymbol V}) = \phi (\prescript{k}{}{{\boldsymbol{x}_N}}) + \sum_{t=0}^{N-1} l (\prescript{k}{}{\boldsymbol{x}_t})$ represents the cost of the $k$-th sample, $\hat{\boldsymbol{u}}_t$ denotes the initial estimate of the control input, and $\tilde{\boldsymbol{u}}_t$ denotes the nominal control input, respectively. $\lambda$ represents the temperature parameter. MPPI utilizes the sampled control inputs to compute a weighted average, allowing it to obtain the optimal control sequence without the need for iterative updates. In practice, only the first control input from the computed sequence is applied to the system. A detailed derivation can be found in \cite{williams2018information}.

\begin{figure*}[t]
    \centering
    \begin{minipage}{0.19\linewidth}
        \subfloat[]{\includegraphics[width=\linewidth]{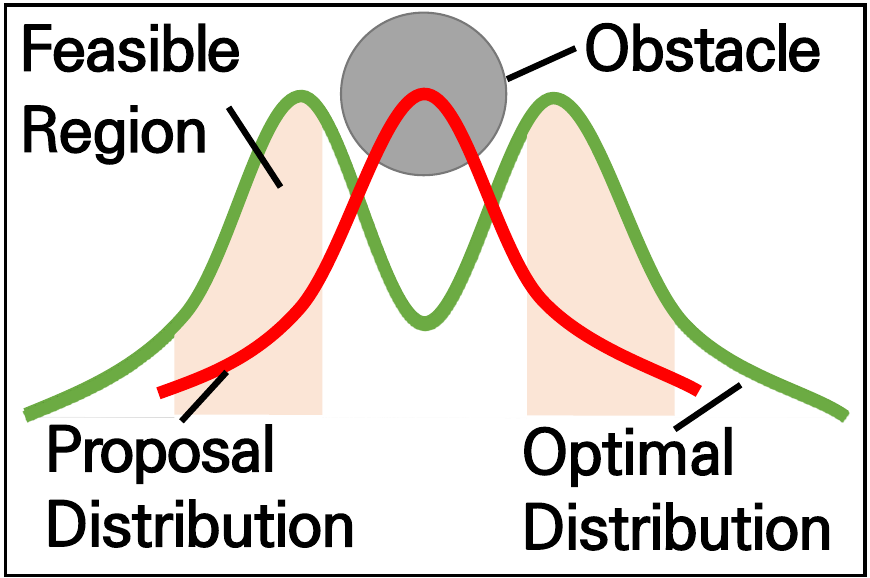} \label{fig2a}}
    \end{minipage}
    \begin{minipage}{0.19\linewidth}
        \subfloat[]{\includegraphics[width=\linewidth]{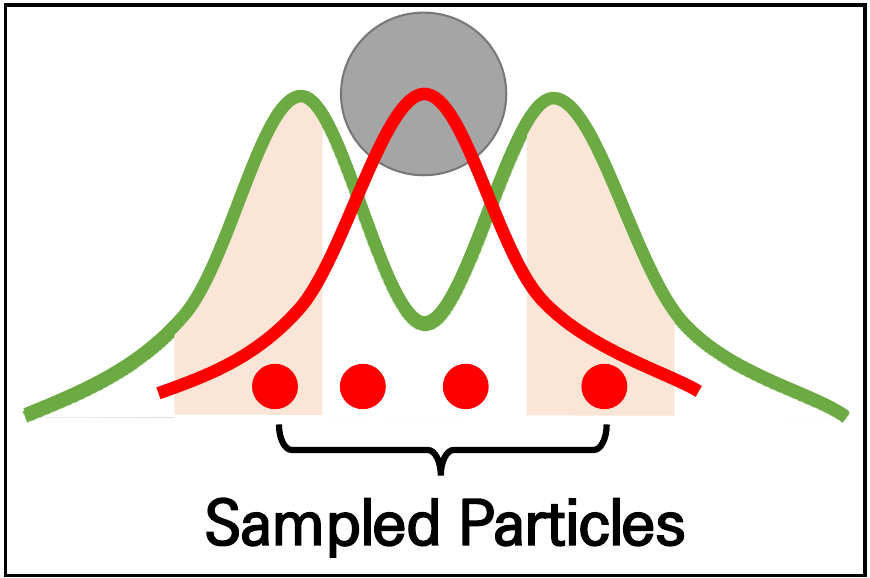} \label{fig2b}}
    \end{minipage}
    \begin{minipage}{0.19\linewidth}
        \subfloat[]{\includegraphics[width=\linewidth]{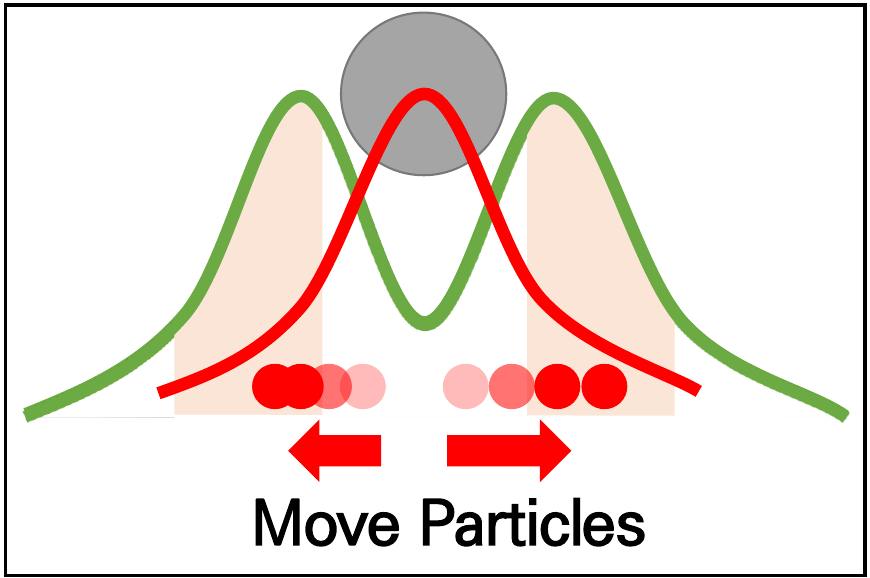} \label{fig2c}}
    \end{minipage}
    \begin{minipage}{0.19\linewidth}
        \subfloat[]{\includegraphics[width=\linewidth]{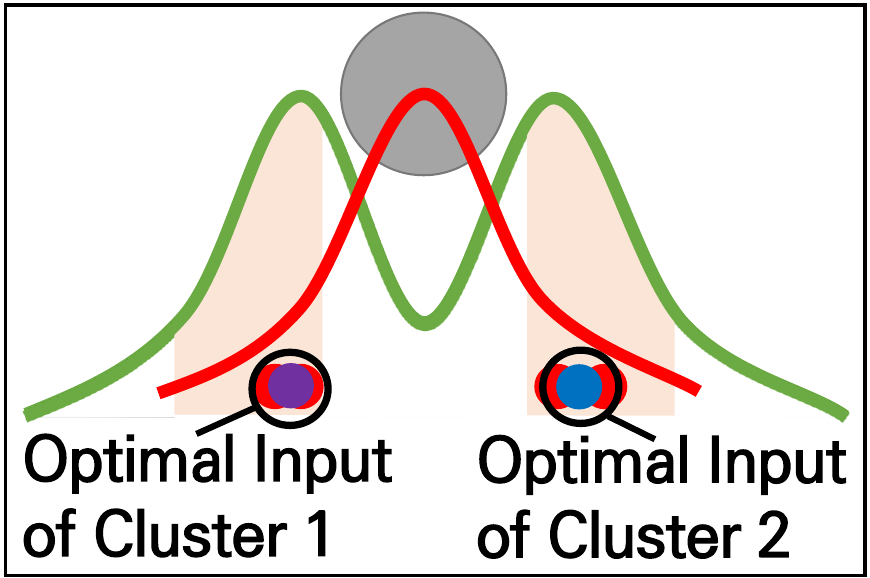} \label{fig2d}}
    \end{minipage}
    \begin{minipage}{0.19\linewidth}
        \subfloat[]{\includegraphics[width=\linewidth]{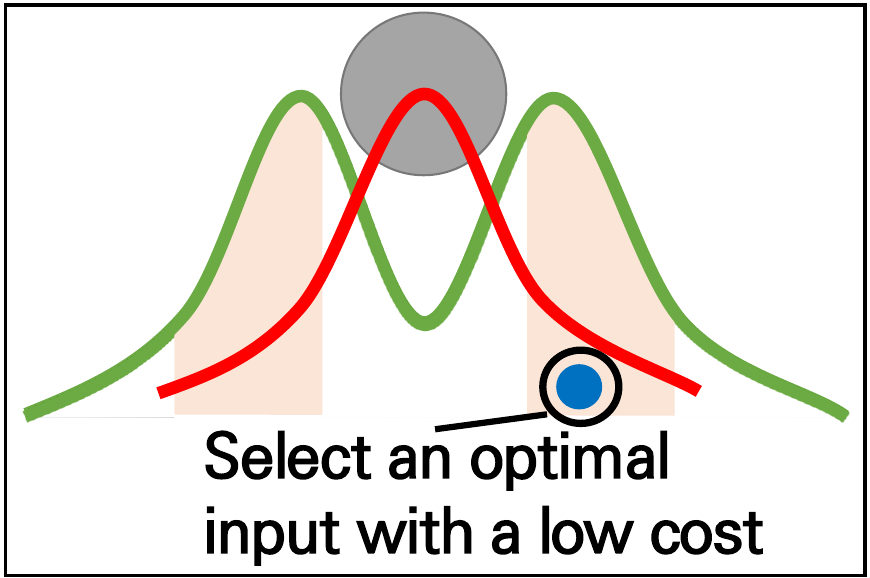} \label{fig2e}}
    \end{minipage}

    \caption{Overview of the CSC-MPPI process.
(a) The gray circle represents an obstacle, while the green and red distributions correspond to the optimal and proposal distributions, respectively. The beige-colored region indicates the feasible region where both state and control input constraints are satisfied.
(b) The sampled particles are randomly drawn from the proposal distribution.
(c) Then, the sampled particles are moved into the feasible region using the primal-dual gradient method.
(d) The sampled particles are clustered using DBSCAN, and the optimal input for each cluster is determined.
(e) Among the optimal inputs from each cluster, the one with the lowest cost is selected as the final optimal input.}
    \label{fig2}
    \vspace{-6mm}
\end{figure*}

\section{CONSTRAINED SAMPLING CLUSTER MPPI} \label{sec3}
As discussed in Section \ref{sec1}, MPPI often struggles to handle hard constraints on states and control inputs. Since trajectory updates in MPPI emerge from stochastic sampling and weighted averaging, naive application can lead to locally suboptimal or even infeasible solutions—particularly when sampled trajectories cluster in high-cost regions. In such scenarios, MPPI may become trapped in local minima, generate suboptimal paths, or increase the risk of collisions. To address these issues, the proposed CSC-MPPI framework combines a primal-dual gradient-based adjustment with DBSCAN clustering.

CSC-MPPI addresses two major challenges in environments with strict constraints:  
(i) ensuring that random samples satisfy strict state and control requirements, and  
(ii) preventing the risk of constraint violations that may arise during weighted averaging when sampled costs are similar.  
By utilizing constraint-aware and density-based clustering, DBSCAN further mitigates potential instability in weighted averaging under such conditions. The proposed approach enhances feasibility, improves robustness, and prevents undesired local minima in complex control tasks.

\subsection{Primal-Dual Gradient-Based Constraint Enforcement} \label{sec3a}
As shown in Fig.~\ref{fig2b} and \ref{fig2c}, the main idea of the proposed method is to shift the randomly sampled candidates into the feasible region defined by the prescribed constraints. Formally, this can be expressed as follows:

\begin{equation} \label{eq_9}
    \begin{aligned}
        &\min_{\prescript{k}{}{\boldsymbol v}_t} {g}(\prescript{k}{}{\boldsymbol{x}}_{t+1}) \cdot \mathbf{1}_{\{g(\prescript{k}{}{\boldsymbol X}) > 0\}} \\
        &\text{s.t.} \quad 
        \prescript{k}{}{\boldsymbol{x}}_{t+1} = f(\prescript{k}{}{\boldsymbol{x}}_t, \prescript{k}{}{\boldsymbol{v}}_t), \\
        &\phantom{\text{s.t.} \quad \;\;} 
        \underline{\boldsymbol {v}} \leq \prescript{k}{}{\boldsymbol v}_t \leq \overline {\boldsymbol {v}},
    \end{aligned}
\end{equation}
where $g(\prescript{k}{}{\boldsymbol x}_{t+1})$ is a concave inequality constraint function that  indicates whether a trajectory satisfies the required conditions. The notation $\prescript{k}{}{\boldsymbol X} = \{\prescript{k}{}{\boldsymbol x_0}, \prescript{k}{}{\boldsymbol x_1}, ..., \prescript{k}{}{\boldsymbol{x}_N} \}^T$ denotes the state sequence of the $k$-th sampled trajectory over the prediction horizon. The indicator function $\boldsymbol{1}_{\{g(\prescript{k}{}{\boldsymbol X}) > 0\}}$  evaluates to 1 if  $g(\prescript{k}{}{\boldsymbol X}) > 0$, and 0 otherwise, thereby ensuring that any infeasible trajectories (i.e. those with $g(\prescript{k}{}{\boldsymbol X})> 0$) are adjusted during the optimization. Additionally, $\underline{\boldsymbol{v}}$ and $\overline{\boldsymbol{v}}$ represent the lower and upper bounds of the control input, respectively, restricting control inputs to the feasible range.

To solve (\ref{eq_9}), we employ a primal-dual gradient method, an iterative strategy widely used for constrained optimization. In this framework, the primal variable (the optimization variable $\prescript{k}{}{\boldsymbol{v}_t}$) is updated via the gradient of the Lagrangian, where the dual variables (Lagrange multipliers) are updated to enforce satisfaction of the constraints. Through this iterative process, the optimization objective is aligned with constraint satisfaction, ensuring convergence to a feasible and optimal solution. The Lagrangian function at time step $t$ for the $k$-th sampled candidate is defined as follows:

\begin{equation} \label{eq_10}
    \begin{aligned}
        \prescript{k}{}{\mathcal{L}}_t(\prescript{k}{}{\boldsymbol{v}}_t, \!\prescript{k}{}{\underline{\boldsymbol{\mu}}}_{t}, \!\prescript{k}{}{\overline{\boldsymbol{\mu}}}&_{t}) = {g}(\prescript{k}{}{\boldsymbol{x}}_{t+1}) \cdot \mathbf{1}_{\{g(\prescript{k}{}{\boldsymbol X}) > 0\}}  \\
        &\quad + \!\prescript{k}{}{\underline{\boldsymbol{\mu}}}_{t}^T (\underline{\boldsymbol{v}} - \prescript{k}{}{\boldsymbol{v}}_t) + \!\prescript{k}{}{\overline{\boldsymbol{\mu}}}_{t}^T (\prescript{k}{}{\boldsymbol{v}}_t - \overline{\boldsymbol{v}}),
    \end{aligned}
\end{equation}
where  $\prescript{k}{} {\underline{\boldsymbol \mu}_t}$ and $\prescript{k}{}{\overline{\boldsymbol \mu}_t}$ are the Lagrange multipliers corresponding to the lower and upper control input bounds, respectively. The gradient of the Lagrangian with respect to $\prescript{k}{}{\boldsymbol{v}_t}$ is then given by:

\begin{equation} \label{eq_11}
    \begin{aligned}
        {\nabla}_{ \prescript{k}{}{\boldsymbol v}_t} \prescript{k}{}{\mathcal{L}}_t(\prescript{k}{}{\boldsymbol{v}}_t, \prescript{k}{}{\underline{\boldsymbol{\mu}}}_{t},\prescript{k}{}{\overline{\boldsymbol{\mu}}}_{t})= 
         {\nabla}_{\prescript{k}{} {\boldsymbol v}_t} g(\prescript{k}{}{\boldsymbol x}_{t+1}) &\cdot \mathbf{1}_{\{g(\prescript{k}{} {\boldsymbol  X}) > 0\}}\\ - \prescript{k}{}{\underline{\boldsymbol{\mu}}}_{t}
        + \prescript{k}{}{\overline{\boldsymbol{\mu}}}_{t},
    \end{aligned}
\end{equation}
where
\begin{equation} \label{eq_12}
    {\nabla}_{\prescript{k}{}{\boldsymbol v}_t} {g}(\prescript{k}{}{\boldsymbol{x}}_{t+1}) = 
    \frac{\partial {g}(\prescript{k}{}{\boldsymbol{x}}_{t+1})}
    {\partial {f}(\prescript{k}{}{\boldsymbol{x}}_t, \prescript{k}{}{\boldsymbol{v}}_t)}
    \frac{\partial {f}(\prescript{k}{}{\boldsymbol{x}}_t, \prescript{k}{}{\boldsymbol{v}}_t)}
    {\partial \prescript{k}{}{\boldsymbol{v}}_t}.
\end{equation}
To maintain feasibility, the Lagrange multipliers are updated by projecting them onto the nonnegative orthant, thus preventing negative values that would violate the Karush-Kuhn-Tucker (KKT) conditions \cite{li2018generalized}:
\begin{equation} \label{eq_13}
    \begin{aligned}
        \prescript{k}{}{\underline{\boldsymbol{\mu}}}_t^{\text{new}} &= \max(0, \prescript{k}{}{\underline{\boldsymbol{\mu}}}_t + \boldsymbol{\beta}_1 \circ (\underline{\boldsymbol{v}} - \prescript{k}{}{\boldsymbol{v}_t})) \\
        \prescript{k}{}{\overline{\boldsymbol{\mu}}}_t^{\text{new}} &= \max(0, \prescript{k}{}{\overline{\boldsymbol{\mu}}}_t + \boldsymbol{\beta}_2 \circ (\prescript{k}{}{\boldsymbol{v}_t} - \overline{\boldsymbol{v}})),
    \end{aligned}
\end{equation}
where $\boldsymbol \beta_1$ and $\boldsymbol \beta_2$ are step size parameters controlling how aggressively each multiplier is adjusted, and $\circ$ denotes the element-wise product operator. The primal variable $\prescript{k}{} {\boldsymbol{v}_t}$ is then updated via gradient descent as follows:
\begin{equation} \label{eq_14}
    \prescript{k}{}{\boldsymbol{v}}^\text{new}_t = \prescript{k}{}{\boldsymbol{v}}_t - \boldsymbol{\alpha} \circ {\nabla}_{\prescript{k}{}{\boldsymbol{v}}_t} \prescript{k}{}{\mathcal{L}}_t(\prescript{k}{}{\boldsymbol{v}_t}, \prescript{k}{}{\underline{\boldsymbol{\mu}}}_{t}, \prescript{k}{}{\overline{\boldsymbol{\mu}}}_{t}),
\end{equation}
where $\boldsymbol \alpha$ is a step size that determines the magnitude of the update. This approach is designed to enforce the KKT conditions, ensuring feasibility in constrained optimization problems. The algorithm alternates between updating the primal variable $\prescript{k}{}{\boldsymbol{v}_t}$ and the dual variables $\prescript{k}{}{\underline{\boldsymbol\mu}_t}$ and $\prescript{k}{}{\overline{\boldsymbol\mu}_t}$. Iterations continue until all KKT conditions are satisfied, ensuring convergence to an optimal solution. Specifically, the algorithm iterates until primal feasibility, dual feasibility, stationarity, and complementary slackness conditions hold. By iteratively re-adjusting control inputs and multiplier values, this primal-dual gradient method enforces the constraints throughout the optimization process and ensures that infeasible trajectories are pushed into the feasible set.

\subsection{Clustering Sampled Trajectories Using DBSCAN} \label{sec3b}
Even though the primal-dual gradient step ensures that the sampled trajectories remain within the feasible region, the weighted averaging process has the potential to produce an optimal trajectory that violates the constraints. To prevent this, we utilize DBSCAN, a density‐based clustering technique, into the trajectory selection process. Unlike traditional clustering algorithms such as k-means, which require a predefined number of clusters, DBSCAN autonomously adapts to both the spatial distribution and cost variation of the samples. Instead of imposing a fixed cluster structure, DBSCAN identifies groups based on density, allowing it to dynamically adjust to the sampled trajectories. It classifies samples into core points, which have sufficiently many neighbors within a specified radius; border points, which lie near core points but lack enough neighbors to be core themselves; and outliers (noise points), which do not fit into any cluster. This density-based approach is particularly advantageous for MPPI, as the number and configuration of feasible trajectories vary due to the stochastic nature of random sampling in each iteration. By filtering out outliers, DBSCAN reduces the likelihood of incorporating unstable or unreliable control inputs in subsequent averaging steps, thereby improving the robustness of the control process.

Several studies have explored the integration of DBSCAN with MPPI for trajectory selection \cite{patrick2024path, jung2024bic}. However, these approaches primarily focus on improving sampling efficiency rather than explicitly addressing constraint violations caused by weighted trajectory averaging. In contrast, the proposed framework leverages DBSCAN to enforce constraint-aware trajectory selection, ensuring that the final control input remains feasible. Specifically, each sampled control input noise $\prescript{k}{}{ \boldsymbol \delta \boldsymbol U}$ is paired with its corresponding cost $S(\prescript{k}{}{\boldsymbol V})$ to form the input data for DBSCAN. The algorithm then groups these control inputs by considering both cost similarity and trajectory geometry similarity. Within each cluster, a cost-weighted average control input is computed following the usual MPPI procedure, as shown in Fig. \ref{fig2d}. Finally, among the averaged control inputs generated from each cluster, the one with the lowest cost is selected for execution, as depicted in Fig. \ref{fig2e}. By filtering out high-cost or inconsistent trajectories before averaging, DBSCAN promotes feasible solutions, mitigating the tendency to drift toward suboptimal or constraint-violating directions. The integration of DBSCAN not only improves the resilience of MPPI to local minima and outlier samples but also enforces feasibility under the most stringent state and control constraints.

Algorithm~\ref{alg:clustered_mppi} summarizes the overall CSC-MPPI pipeline. The initial sampled trajectories are generated via standard MPPI and propagated using the system dynamics (Lines 2–7). A primal-dual gradient update is then applied to each sample to satisfy state and control constraints (Lines 8–15). After computing the total cost of each trajectory (Lines 16–20), the control deviations and corresponding costs are stored as a dataset $\mathcal{D} = \{(\delta \boldsymbol{U}^k, S^k)\}$, which serves as input to DBSCAN (Lines 22-23). This clustering step groups the samples based on control similarity and cost proximity, while automatically discarding outliers. A softmax-weighted MPPI update is performed on each cluster (Lines 24–26), using only the samples within that cluster. Here, $\rho$ denotes the minimum cost within the cluster and serves as a reference to stabilize the weight computation. Finally, the cluster with the lowest-cost result is selected to generate the control output (Line 27).

\begin{algorithm}[t]
\caption{Constrained Sampling Cluster MPPI}
\label{alg:clustered_mppi}
\begin{algorithmic}[1]
\Require 
 \Statex$f, \boldsymbol \Sigma_{\boldsymbol u},\lambda$: {Dynamics, Noise Covariance, Temperature}
 \Statex{$x_0, \boldsymbol U$: {Initial Condition, Initial Control Sequence}}
 \Statex{$K, N$: Number of Samples, Number of Time Steps}
 \Statex{$\boldsymbol{\alpha}, \boldsymbol{\beta}_1, \boldsymbol{\beta}_2$}: Primal-Dual Gradient Step Size
 \Statex{$l,\phi$: Running Cost, Terminal Cost}

    \For{$k \gets 0$ to $K-1$} \Comment{In Parallel}
        \State $\prescript{k}{}{\boldsymbol{x}}_0 \gets \boldsymbol x_0$
        \For{$t \gets 0$ to $N - 1$}
            \State $\delta \prescript{k}{}{\boldsymbol{u}}_t \sim \mathcal{N}(0, \boldsymbol \Sigma_{\boldsymbol u})$
            \State $\prescript{k}{}{\boldsymbol{v}}_t \gets \boldsymbol u_t + \delta \prescript{k}{}{\boldsymbol{u}}_t$
            \State $\prescript{k}{}{\boldsymbol{x}}_{t+1} \gets f(\prescript{k}{}{\boldsymbol{x}}_t, \prescript{k}{}{\boldsymbol{v}}_t)$
        \EndFor
        \While{\text{ KKT Conditions not Satisfied}}
            \For{$t \gets 0$ to $N - 1$}
                \State $\prescript{k}{}{\underline {\boldsymbol{\mu}}}_t \gets \max(0, \prescript{k}{}{\underline{\boldsymbol{\mu}}}_t + \boldsymbol \beta_1 \circ (\underline{\boldsymbol v} - \prescript{k}{}{\boldsymbol{v}}_t))$
                \State $\prescript{k}{}{\overline{\boldsymbol{\mu}}}_t \gets \max(0, \prescript{k}{}{\overline{\boldsymbol{\mu}}}_t + \boldsymbol \beta_2 \circ(\prescript{k}{}{\boldsymbol{v}}_t - \overline {\boldsymbol v}) )$
                \State $\prescript{k}{}{\boldsymbol{v}}_t \gets \prescript{k}{}{\boldsymbol{v}}_t -  \boldsymbol{\alpha} \circ{\nabla}_{\prescript{k}{}{\boldsymbol{v}_t}} \prescript{k}{}{\mathcal{L}}_t(\prescript{k}{}{\boldsymbol{v}_t}, \prescript{k}{}{\underline{\boldsymbol{\mu}}}_{t},\prescript{k}{}{\overline{\boldsymbol{\mu}}}_{t})$
                
                \State $\prescript{k}{}{\boldsymbol{x}}_{t+1} \gets f(\prescript{k}{}{\boldsymbol{x}}_t, \prescript{k}{}{\boldsymbol{v}}_t)$
            \EndFor
        \EndWhile

        \For{$t \gets 0$ to $N - 1$}
            \State $\delta \prescript{k}{}{\boldsymbol{u}}_t \gets \prescript{k}{}{\boldsymbol{v}}_t-\boldsymbol u_t$
            \State $\prescript{k}{}{S} \gets \prescript{k}{}{S} + l(\prescript{k}{}{\boldsymbol{x}}_t) +  \boldsymbol{u}_t^T \boldsymbol \Sigma_{\boldsymbol u}^{-1} \prescript{k}{}{\boldsymbol{v}}_t$
        \EndFor
        
        \State $\prescript{k}{}{S} \gets \prescript{k}{}{S} + \phi(\prescript{k}{}{\boldsymbol{x}}_N)$
        
    \EndFor
    \State $\mathcal{D} \gets \{ (\prescript{k}{}{\delta \boldsymbol U}, \prescript{k}{}{S}) \mid k = 0, \ldots, K-1 \}$
    
    \State $\{C_0, \ldots, C_{M-1}\} \gets \text{DBSCAN}(\mathcal{D})$
    
    \For{$m \gets 0$ to $M-1$}
        \State $\boldsymbol U^*_{C_m} \gets \text{MPPI}(\lambda, \boldsymbol{U}, \boldsymbol{\delta U}_{C_m}, S_{C_m})$
    \EndFor
    
    \State $\boldsymbol U^* \gets \arg\min_{\boldsymbol U^*_{C}} S(\boldsymbol U^*_{C})$ 
    
    \Function{MPPI}{$\lambda, \boldsymbol U, \boldsymbol{\delta U},S$}
        \State $\rho \gets \min(\prescript{0}{}{S}, \cdots, \prescript{d-1}{}{S})$
         \State $\eta \gets \sum_{i=0}^{d-1} \exp\left(-\frac{1}{\lambda}(\prescript{i}{}{S} - \rho)\right)$
        \For{$i \gets 0$ to $d-1$}
            \State $\prescript{i}{}{w} \gets \frac{1}{\eta} \exp\left(-\frac{1}{\lambda}(\prescript{i}{}{S} - \rho)\right)$
        \EndFor
        \State $\boldsymbol U^* \gets \boldsymbol U + \sum_{i=0}^{d-1} \prescript{i}{}{w} \delta \prescript{i}{}{\boldsymbol U}$
        
        \State \Return $\boldsymbol U^*$
    \EndFunction
    
\end{algorithmic}
\end{algorithm}

\section{EXPERIMENTS} \label{sec4}

In this section, we validate the effectiveness of the proposed method through simulations and real-world experiments. In the simulation environment, we compare the proposed method with standard MPPI in an obstacle avoidance task. Additionally, to examine the risk of constraint violation due to weighted averaging, we evaluate the impact of DBSCAN-based clustering by comparing a full version of the proposed CSC-MPPI and a simplified variant that omits the DBSCAN clustering step. Finally, the real-world experiments were conducted to validate CSC-MPPI in practical scenarios.

\subsection{Simulation Scenarios} \label{sec4A}

In both simulation environments shown in Fig.~\ref{fig3}, the robot starts from the initial state indicated by a black circle and navigates toward the goal state marked by a purple circle. Obstacles are represented by gray circles, and the dashed boundaries around them reflect safety margins that account for the robot's footprint.

As shown in Fig.~\ref{fig3a}, \textit{Environment 1} is designed to compare the performance of standard MPPI and CSC-MPPI. The autonomous robot starts from 
\(\boldsymbol{x}_s = [-1.0, -1.0, \pi/2]^T\) and must reach 
\(\boldsymbol{x}_f = [2.0, 2.0, \pi/2]^T\) while avoiding one dynamic and two static obstacles. The dynamic obstacle has a radius of 0.3 m and moves from \([-1.0, 0.0]^T\) to \([0.5, 0.0]^T\) at a constant velocity of 0.53 m/s. The static obstacles are centered at \([0.0, 1.0]^T\) and \([1.5, 0.7]^T\), with radii of 0.4 m and 0.5 m, respectively.

As shown in Fig.~\ref{fig3b}, \textit{Environment 2} is a simpler scenario used to analyze the performance of CSC-MPPI and its variant without DBSCAN. The robot begins at 
\(\boldsymbol{x}_s = [-1.0, 0.0, 0.0]^T\) and aims to reach 
\(\boldsymbol{x}_f = [1.0, 0.0, 0.0]^T\), navigating around a single static obstacle centered at \([0.0, 0.0]^T\) with a radius of 0.5 m.

\subsection{Simulation Setup} \label{sec4B}
In this study, we employ a differential-drive robot and utilize its kinematic model. The system dynamics are given as follows:

\begin{figure}[t] 
    \centering
    \subfloat[]{\includegraphics[width=0.45\linewidth]{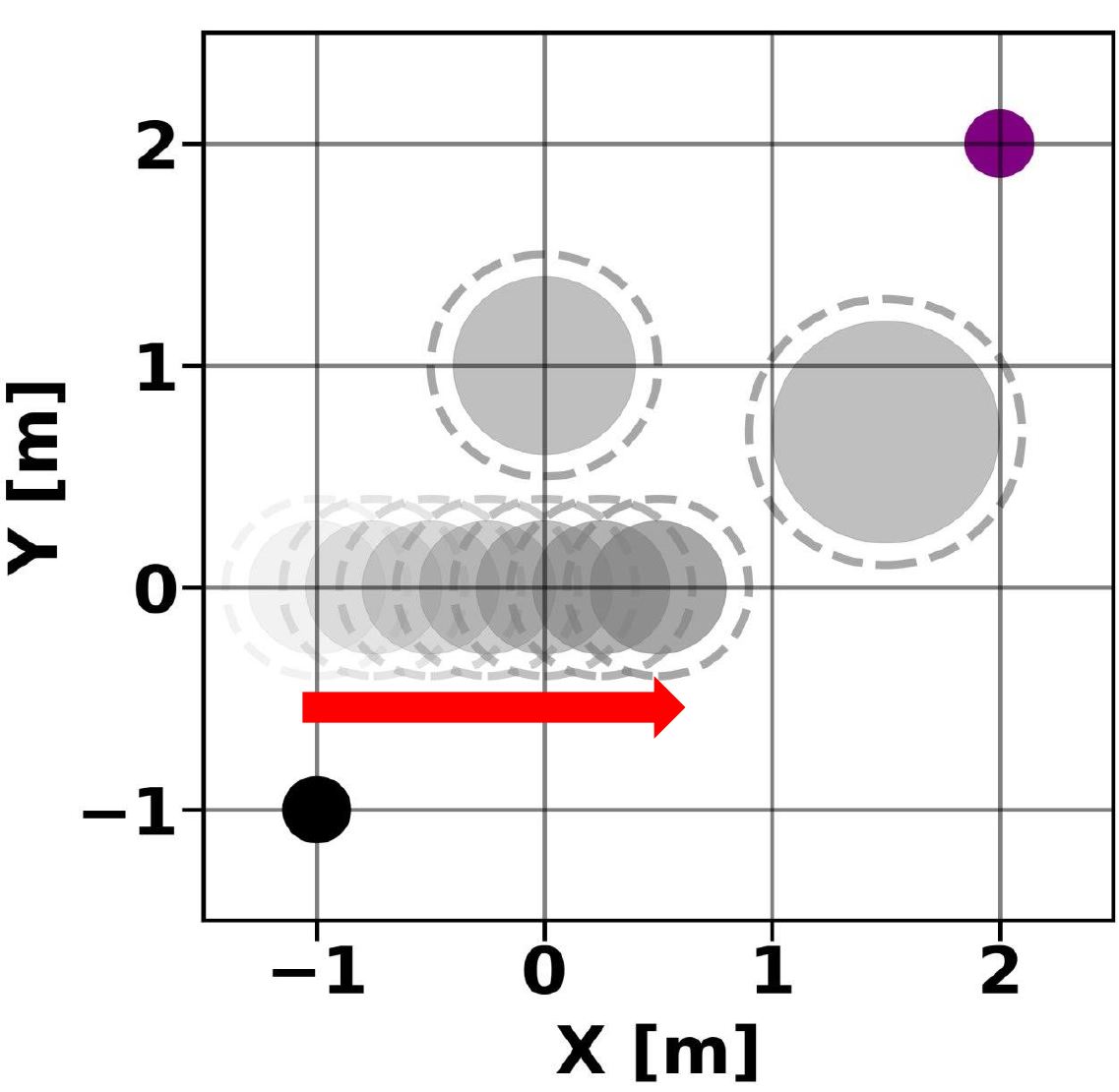} \label{fig3a}}
    \subfloat[]{\includegraphics[width=0.45\linewidth]{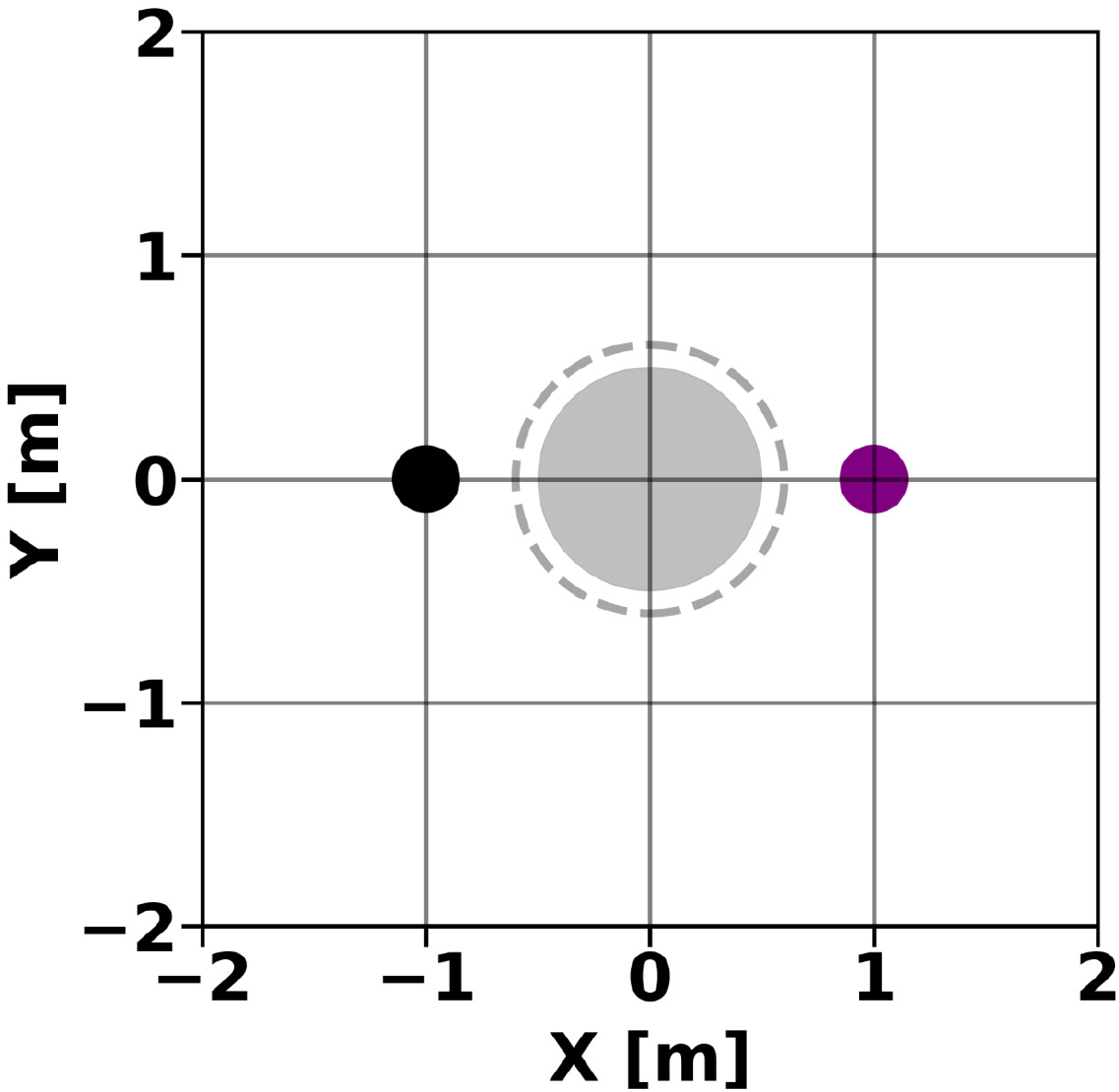} \label{fig3b}} \\

    \caption{Simulation environments used for performance evaluation. (a) \textit{Environment \#1} for comparison between standard MPPI and CSC-MPPI. (b) \textit{Environment \#2} for analyzing the impact of DBSCAN.
}
    \label{fig3}
    \vspace{-6mm}
\end{figure}

\begin{equation} \label{eq_15}
    \begin{split}
        x_{t+1} &= x_t + u_{v_t} \cos\theta_t \, dt \\
        y_{t+1} &= y_t + u_{v_t} \sin\theta_t \, dt \\
        \theta_{t+1} &= \theta_t + u_{\omega_t} \, dt.
    \end{split}
\end{equation}
The system state $\boldsymbol{x} = [x,y,\theta]^T$ represents the position and orientation of the robot, while the control input $\boldsymbol{u} = [u_v, u_w]^T$ represents the linear and angular velocity commands. The simulations were conducted on a computer equipped with an Intel Core i7-13700F processor running at 2.1 GHz, 32 GB of RAM, and an NVIDIA GeForce RTX 4060 GPU. Parallel computation was performed using a Python-based simulation framework. For simulation Environment \textit{\#1}, the temperature parameter $\lambda$ was set to 0.01, while for Environment \textit{\#2}, $\lambda$ was set to 0.7. This higher temperature in Environment \textit{\#2} was deliberately chosen to highlight the effect of DBSCAN within the CSC-MPPI framework. In Environment \textit{\#2}, the number of sampled trajectories was fixed at $K = 300$. The remaining MPPI parameters were kept identical across both environments. The covariance matrix $\boldsymbol \Sigma_{\boldsymbol u}$ was defined as $\text{diag}(\sigma_{u_v}^2, \sigma_{u_w}^2) = \text{diag}(0.1^2, 1.0^2)$. The time horizon $N$ was set to 30 with a time step of $dt = 0.03$. The linear velocity was constrained between $0$ m/s and $0.5$ m/s, while the angular velocity was constrained between $-3.0$ rad/s and $3.0$ rad/s. The tolerance for the goal state is defined as \(\tau_p = 0.15\) m for the position \((x, y)\) and \(\tau_\theta = 0.25\) rad for the orientation \(\theta\). The running cost \( l \) is defined as  
\begin{equation} \label{eq_16}
    l = (\boldsymbol{x}_t - \boldsymbol{x}_f)^T \boldsymbol{Q} (\boldsymbol{x}_t - \boldsymbol{x}_f),    
\end{equation}
where \(\boldsymbol{Q} = \text{diag}(10,10,0)\). The terminal cost \( \phi \) is defined as  
\begin{equation} \label{eq_17}
    \phi = (\boldsymbol{x}_N - \boldsymbol{x}_f)^T \boldsymbol{H} (\boldsymbol{x}_N - \boldsymbol{x}_f),    
\end{equation}
where \(\boldsymbol{H} = \text{diag}(50,50,50)\).
In the case of CSC-MPPI, the obstacle constraint function is defined as:
\begin{equation} \label{eq_18} g(x) = r^2 - (x - x_{\text{obs}})^2 - (y - y_{\text{obs}})^2, \end{equation}
where $x_{\text{obs}}$ and $y_{\text{obs}}$ are the coordinates of the obstacle center, and $r$ is the radius of the obstacle.
In standard MPPI, a penalty cost is introduced as a soft constraint for obstacle avoidance:
\begin{equation} \label{eq_19}
    C =
\begin{cases}
    10^4, & \text{if collision}, \\
    0, & \text{otherwise}.
\end{cases}    
\end{equation}
Each case was repeated {20 times} in \textit{Environment \#1} and {10 times} in \textit{Environment \#2}.


\begin{figure}[t] \label{fig4}
    \centering
    
    \subfloat[]{\includegraphics[width=0.45\linewidth]{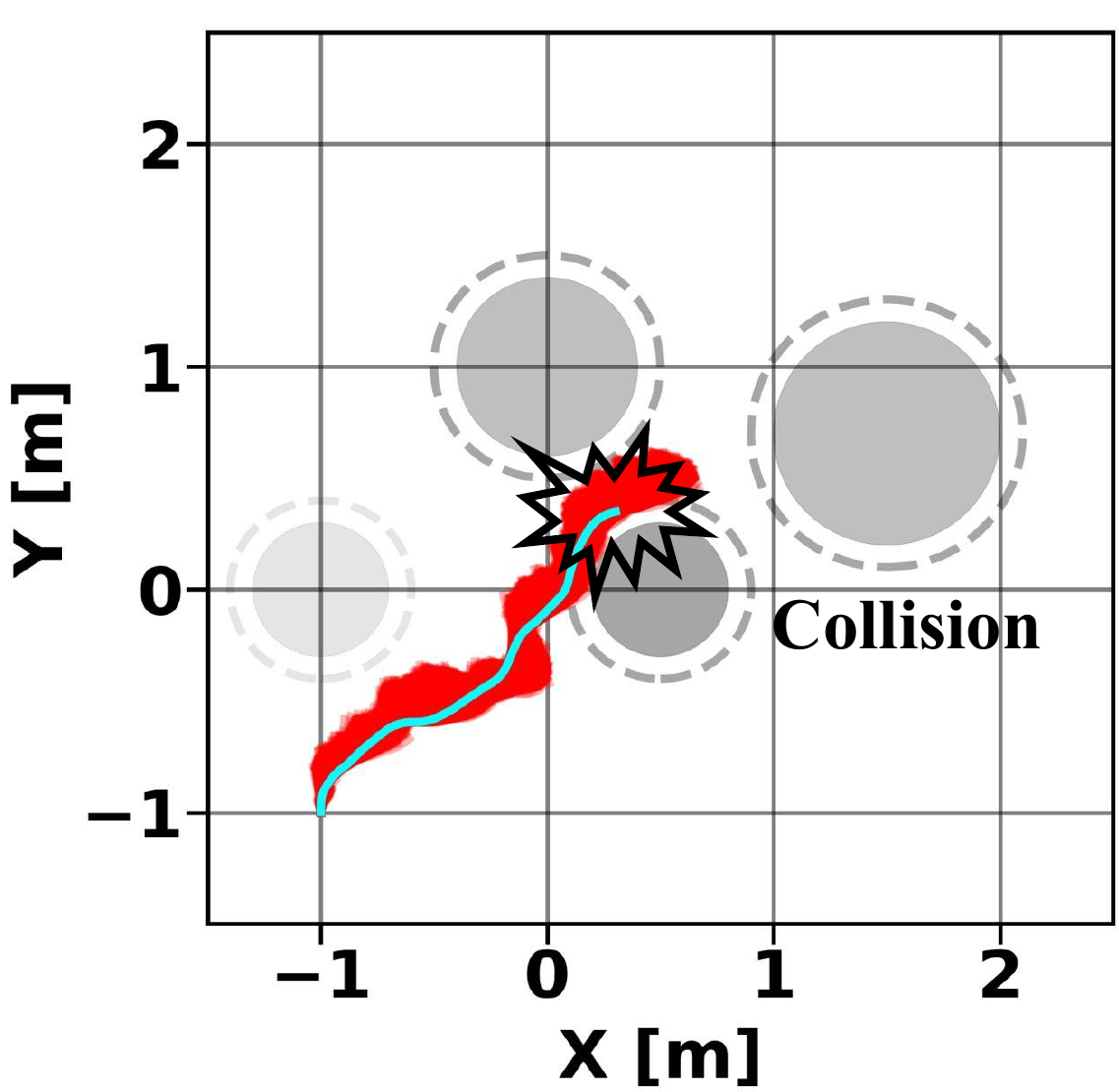} \label{fig4a}}
    \hfill
    \subfloat[]{\includegraphics[width=0.45\linewidth]{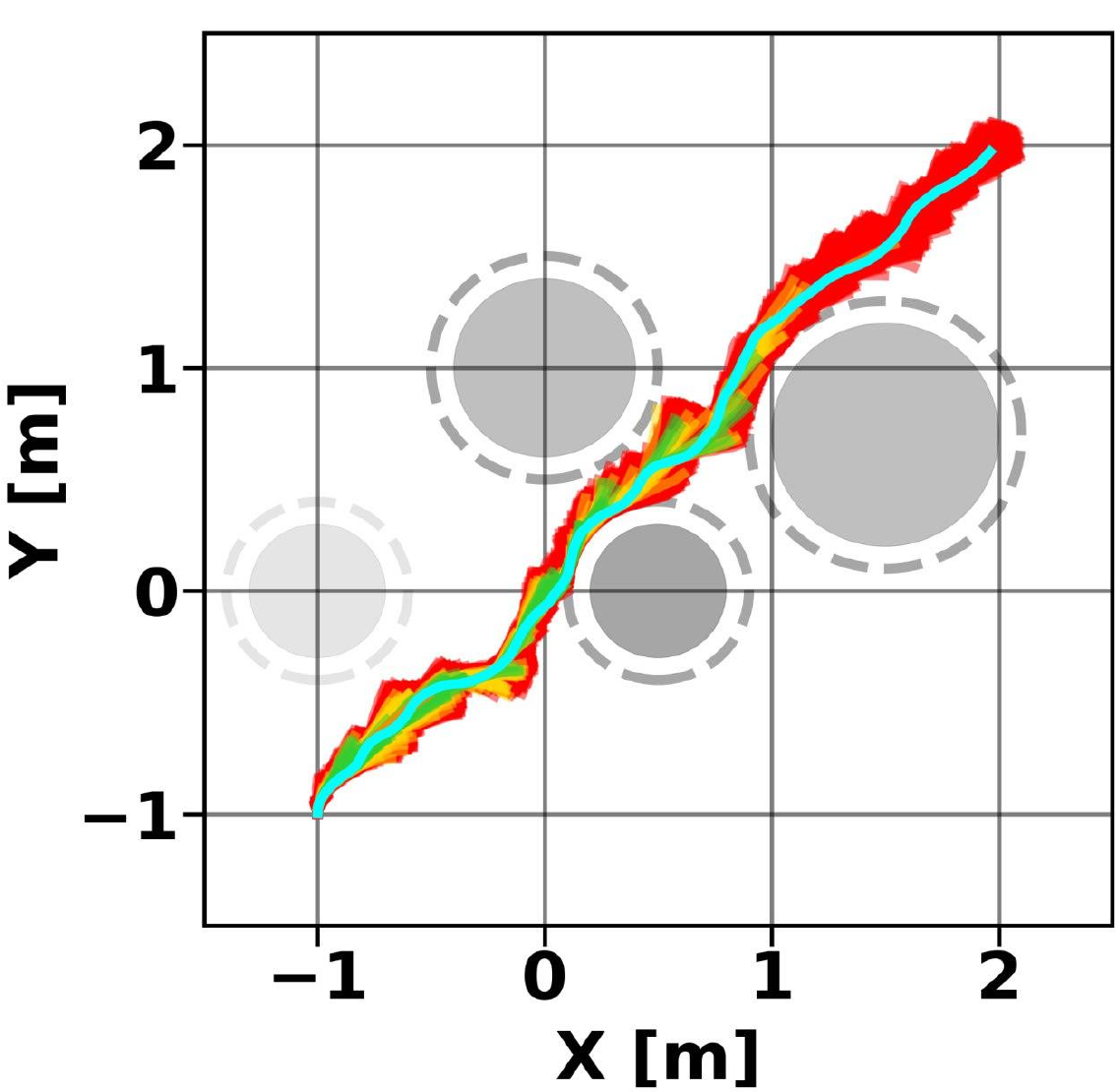} \label{fig4b}}\\

    \caption{Trajectory comparison of standard MPPI and CSC-MPPI for the navigation task in \textit{Environment {\#1}}. (a) The trajectories generated by standard 
    MPPI with soft constraints ($K=300$). The red lines represent sampled trajectories, while the cyan line indicates the optimal path. (b) The trajectories generated by CSC-MPPI ($K=300$). The red, yellow, orange, and green lines represent clustered trajectories, while the cyan line indicates the optimal path.
}
\vspace{-3mm}
\end{figure}

\subsection{Simulation Results}

The results for \textit{Environment {\#1}} are summarized in Table~\ref{table2}. The collision rate represents the proportion of trials in which the robot made contact with an obstacle during 20 test runs. The path length denotes the total distance traveled by the robot to reach the goal, while the average and maximum computation times correspond to the processing time per iteration of the algorithm. These metrics—path length, average computation time, and maximum computation time—are computed exclusively for successful trials.

The evaluation of standard MPPI under different trajectory sample sizes yielded collision rates of 80\%, 50\%, and 30\% for $K=20$, $50$, and $300$, respectively. A reduction in the number of sampled trajectories resulted in a higher probability of collision, suggesting that standard MPPI exhibits reduced robustness in obstacle avoidance when fewer samples are available. In contrast, CSC-MPPI consistently achieved a 0\% collision rate across all scenarios, highlighting the effectiveness of constraint enforcement. The experimental results demonstrate that CSC-MPPI avoids collisions by strictly adhering to feasibility constraints during trajectory optimization. In terms of path length, CSC-MPPI produced the shortest trajectory for $K=300$, with a recorded path length of 4.476 m. Standard MPPI, by comparison, generated longer paths regardless of the number of sampled trajectories. Notably, even with only $K=20$, CSC-MPPI achieved a path length of 4.766 m, which remained shorter than the 4.848 m path generated by standard MPPI at $K=300$.  Since CSC-MPPI explicitly enforces obstacle avoidance as a hard constraint, the optimization focuses on minimizing the cost of reaching the goal while remaining within the feasible solution space. This constraint-driven optimization tends to produce more direct and shorter trajectories.
\begin{table}[t]
    \centering
    \renewcommand{\arraystretch}{1.1} 
    \setlength{\tabcolsep}{7pt} 
    \caption{Performance comparison of standard MPPI and CSC-MPPI with different sample sizes}
    \resizebox{0.49\textwidth}{!}{ 
    \large
    \begin{tabular}{lccccc}
        \hline
        Method &  \makecell{Collision \\ Rate [\%]} & \makecell{Path \\ Length [m]} & \makecell{Average \\ Time [ms]} & \makecell{Max \\ Time [ms]} \\
        \hline
        \hline
        Standard MPPI (K=20)   & 80 & 4.822 & 1.739 & 5.318 \\
        Standard MPPI (K=50)   & 50 & 4.755 & 1.720 & 4.548 \\
        Standard MPPI (K=300)  & 30 & 4.848 & 1.669 & 4.144 \\

        CSC-MPPI (K=20)  & 0  & 4.766 & 8.870 & 19.82 \\
        CSC-MPPI (K=50)   & 0  & 4.629 & 9.007 & 19.31 \\
        CSC-MPPI (K=300)  & 0  & 4.476 & 8.915 & 20.61 \\
        
        \bottomrule
    \end{tabular}
    }
    \label{table2}
    \vspace{-6mm}
\end{table}
As shown in Fig.~\ref{fig4a}, standard MPPI with soft constraints does not always guarantee obstacle avoidance. When all sampled trajectories collide with obstacles or when the weight assigned to obstacle costs is lower than that of other cost terms, obstacle avoidance is not ensured. In contrast, as depicted in Fig.~\ref{fig4b}, CSC-MPPI ensures that all sampled trajectories satisfy the constraints, regardless of cost.


The computational efficiency of each method was also analyzed. Despite increasing the number of sampled trajectories from 20 to 300, both standard MPPI and CSC-MPPI maintained stable computation times ranging from 1.669 ms to 1.739 ms for standard MPPI and from 8.870 ms to 9.007 ms for CSC-MPPI. This consistent performance is attributed to GPU-based parallel computation, which mitigates the computational burden even as sample size increases. The increased computational time in CSC-MPPI is mainly caused by the primal-dual gradient adjustment and the DBSCAN-based clustering process. Specifically, performing DBSCAN clustering and obtaining representative sequences from the resulting clusters took an average of 6.03 ms for \(K=20\), 6.06 ms for \(K=50\), and 5.99 ms for \(K=300\). Notably, CSC-MPPI achieves superior collision avoidance and trajectory efficiency while utilizing only 10\% of the samples required by standard MPPI.

\begin{figure}[t] \label{fig5}
    \centering

    \subfloat[]{\includegraphics[width=0.45\linewidth]{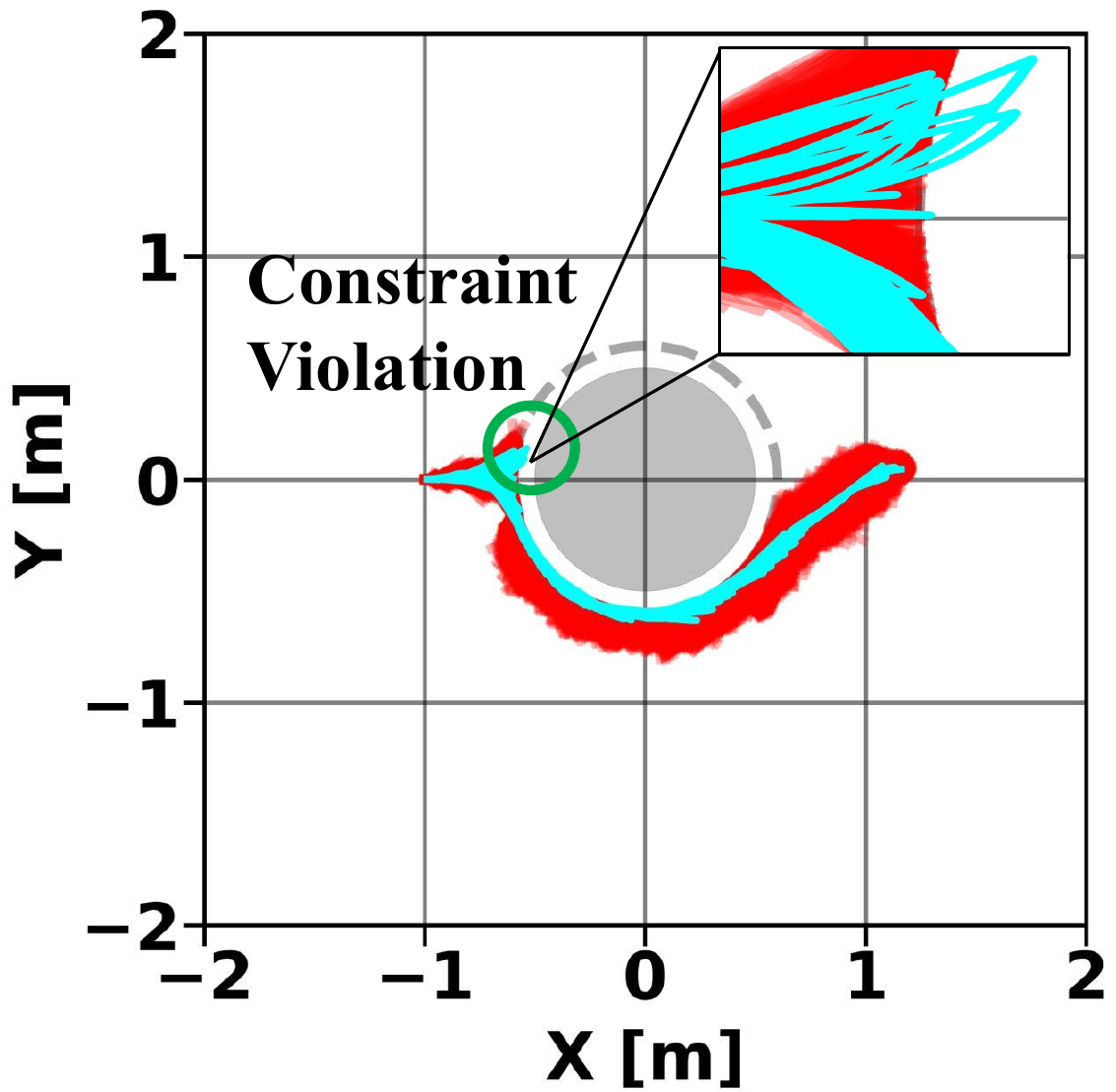} \label{fig5a}}
    \hfill
    \subfloat[]{\includegraphics[width=0.45\linewidth]{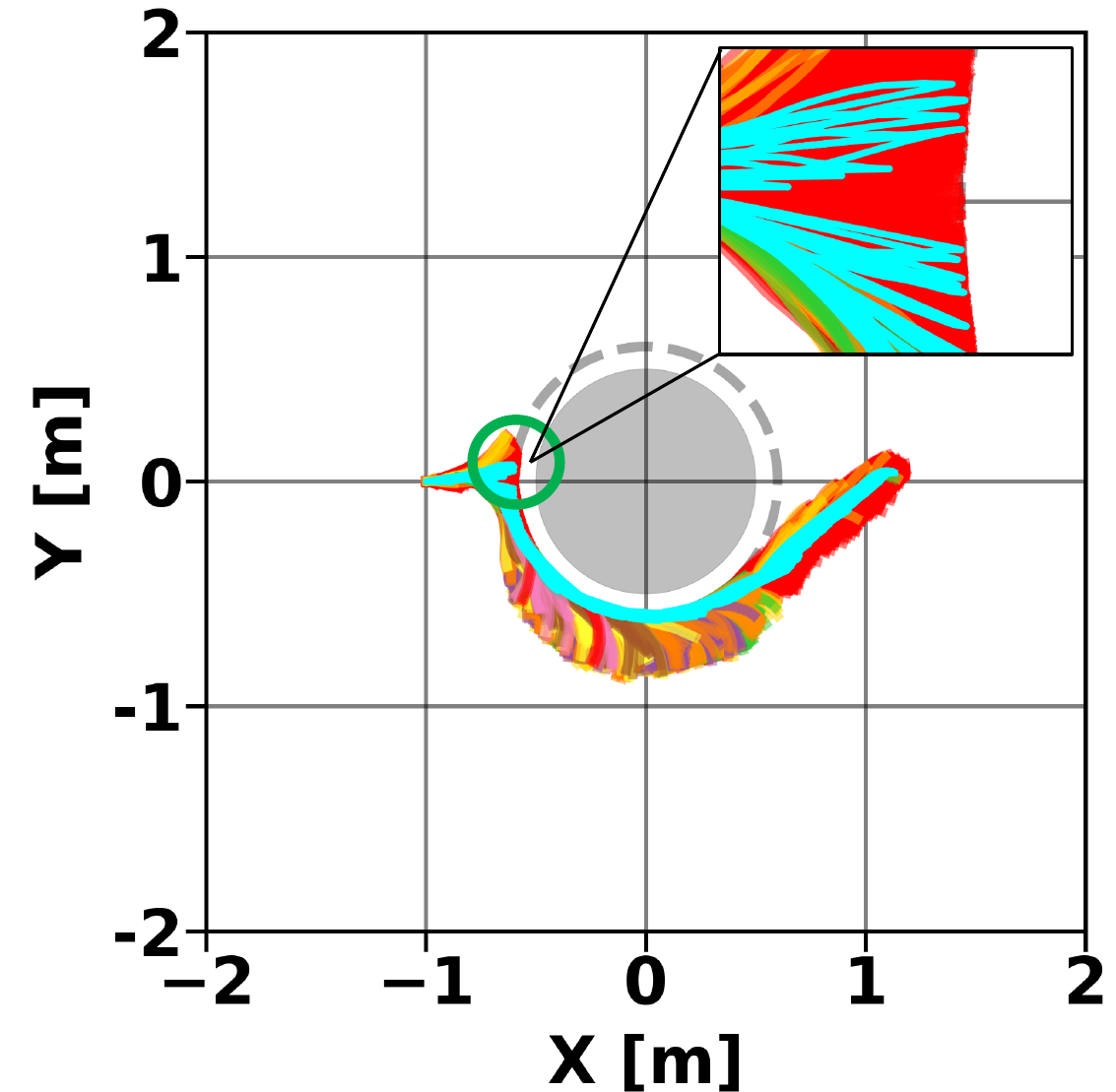} \label{fig5b}}

    \caption{Trajectory comparison between CSC-MPPI and its variant without DBSCAN in \textit{Environment \#2}. (a) The trajectories generated by the CSC-MPPI variant without DBSCAN. The red lines and the cyan line represent the sampled trajectories and the optimal trajectory obtained from (\ref{eq_6}), respectively. (b) The trajectories generated by CSC-MPPI. Colored lines (e.g., red, yellow, orange, pink) represent clustered trajectories, and the cyan line indicates the final optimal trajectory.
}
\vspace{-6mm}
\end{figure}

In \textit{Environment \#2}, the experiment was conducted to evaluate constraint satisfaction by comparing CSC-MPPI and its variant without DBSCAN. Both approaches achieved a {0\% collision rate}. However, when considering constraint satisfaction of the trajectory corresponding to the optimal input sequence {obtained from (\ref{eq_6})}, the satisfaction rate was {100\% for CSC-MPPI}, whereas it dropped to {80\% for the variant without DBSCAN}.
As shown in Fig.~\ref{fig5a}, even though the sampled trajectories satisfy the constraints, if multiple sampled trajectories have similar costs, the weighted averaging process may result in a trajectory that does not satisfy the constraints. In contrast, as illustrated in Fig.~\ref{fig5b}, CSC-MPPI uses DBSCAN to select a representative cluster from the set of feasible trajectories, thereby ensuring that the final optimal input sequence satisfies the constraints. These results demonstrate that the proposed method effectively mitigates the risk of constraint violation introduced by weighted averaging.

\subsection{Real-World Experimental Setup and Results}

\begin{figure}[t]
    \centering
    \includegraphics[width=0.49\textwidth]{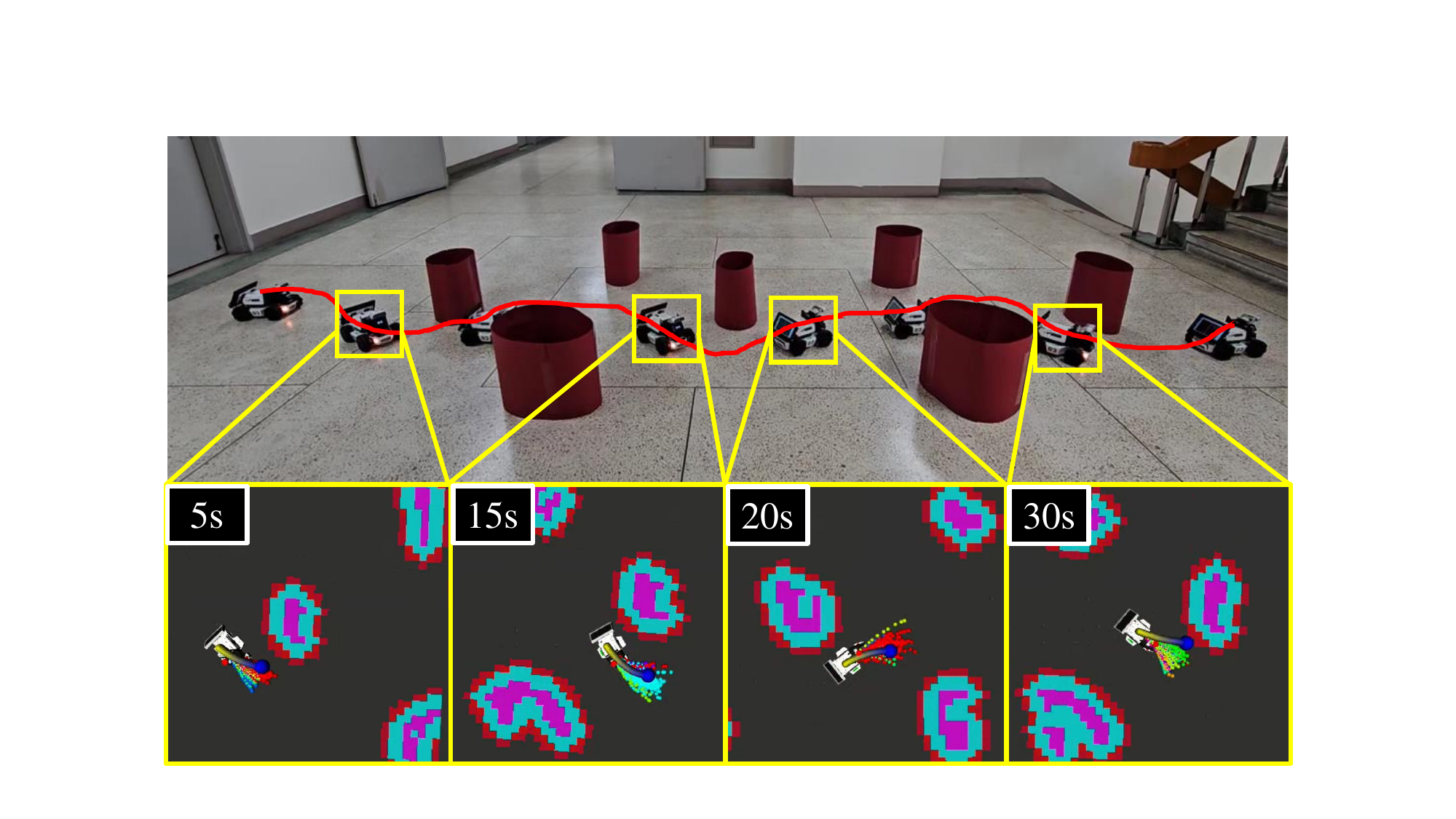}
    \caption{Snapshots of the real-world experiment. The red line and bottom images represent the actual trajectory of the robot and local costmap, respectively.
}
    \label{fig6}
    \vspace{-6mm}
\end{figure}

In addition to the simulation experiments, we conducted real-world experiments using the LIMO robot (\textit{Agile Robotics Co.}). The onboard computing unit was an Intel NUC i7, and the LiDAR sensor used for perception was an EAI T-MINI Pro. Unlike the Python-based simulation, which utilized parallel computation on a GPU, the real-world experiment relied solely on CPU-based computations. For the real-world experiments, the number of sampled trajectories was set to $K=300$, with a time horizon of $N=40$ and a time step of $dt=0.05$. The control frequency was fixed at 10 Hz. The tolerance for the goal state was set as  
$\tau_p = 0.2 \text{ m}$, $\tau_\theta = 0.25 \text{ rad}$. Fig. \ref{fig6} illustrates the real-world experiment, showing the robot's trajectory along with the corresponding local cost map at different time steps.
 The test environment contained a total of seven static obstacles, each with a radius of 0.17 m. The robot was required to navigate from the initial state  $\boldsymbol{x}_s = [0, 0, 0]^T$ to the goal state $\boldsymbol{x}_f = [6.15, 0.3, 0]^T$ while avoiding obstacles. 
 

To evaluate the performance of the proposed approach in real-world scenarios, we conducted 10 repeated trials. The results showed a collision rate of 0\%, demonstrating the robustness of the system in obstacle avoidance. The average computation time was measured at 6.646 ms, with the clustering process accounting for  1.715 ms of the total computation time. The maximum computation time was recorded as 38.89 ms. These results validate the efficiency of the CSC-MPPI framework, highlighting its capability to generate feasible and safe trajectories in real-time applications.

Since the real-world experiments relied on CPU-based computations, we employed a clamping technique in (\ref{eq_9}) to reduce computation time. Clamping is a simple and efficient method that ensures constraint satisfaction by directly clipping control inputs. Given the relatively simple constraints in our setup such as velocity limits, clamping was sufficient without significantly affecting performance. However, in more complex scenarios involving coupled constraints, or high-dimensional constraints, clamping may lead to suboptimal solutions. In such cases, the gradient method proposed in this paper would be more suitable, as it iteratively adjusts dual variables to better handle intricate constraints.



\section{CONCLUSIONS} \label{sec5}
In this paper, we introduced a novel CSC-MPPI framework that integrates a primal-dual gradient-based constraint enforcement method with DBSCAN clustering to overcome the inherent limitations of standard MPPI in satisfying hard constraints. By iteratively adjusting sampled control inputs into the feasible region and clustering them based on spatial and cost similarities, the proposed approach effectively selects the optimal control action while strictly adhering to both state and input constraints. Simulation and real-world experimental results demonstrate that CSC-MPPI achieves a zero collision rate and generates shorter, more feasible trajectories compared to conventional MPPI methods, thereby validating its effectiveness and robustness in complex obstacle avoidance tasks. Future work will focus on enhancing computational efficiency and extending the framework to more challenging environments and diverse robotic platforms including a mobile manipulator and humanoid.

\bibliographystyle{IEEEtran}
\bibliography{IEEEabrv, mine}

\end{document}